\documentclass[conference]{IEEEtran}
\IEEEoverridecommandlockouts

\usepackage{cite}
\usepackage{amsmath,amssymb,amsfonts}
\usepackage{makecell}
\usepackage{graphicx}
\usepackage{textcomp}
\usepackage{xcolor}
\usepackage{xspace}
\usepackage{hyperref}
\usepackage{multirow}
\usepackage{algorithm}
\usepackage{algpseudocode}
\usepackage[normalem]{ulem}
\useunder{\uline}{\ul}{}
\newcommand{\modelname}{Fed-TREND\xspace}

\def\BibTeX{{\rm B\kern-.05em{\sc i\kern-.025em b}\kern-.08em
    T\kern-.1667em\lower.7ex\hbox{E}\kern-.125emX}}
\begin{document}

\makeatletter
\newcommand{\linebreakand}{
  \end{@IEEEauthorhalign}
  \hfill\mbox{}\par
  \mbox{}\hfill\begin{@IEEEauthorhalign}
}
\makeatother

\title{Tackling Data Heterogeneity in Federated Time Series Forecasting}

\author{\IEEEauthorblockN{Wei Yuan}
\IEEEauthorblockA{\textit{The University of Queensland} \\
Brisbane, Australia \\
w.yuan@uq.edu.au}
\and
\IEEEauthorblockN{Guanhua Ye}
\IEEEauthorblockA{\textit{Beijing University of} \\ \textit{Posts and Telecommunications} \\
Beijing, China \\
g.ye@bupt.edu.cn}
\and
\IEEEauthorblockN{Xiangyu Zhao}
\IEEEauthorblockA{\textit{City University of Hong Kong} \\
Hongkong, China \\
xianzhao@cityu.edu.hk}
\linebreakand
\IEEEauthorblockN{Quoc Viet Hung Nguyen}
\IEEEauthorblockA{\textit{Griffith University} \\
Gold Coast, Australia \\
henry.nguyen@griffith.edu.au}
\and
\IEEEauthorblockN{Yang Cao}
\IEEEauthorblockA{\textit{Institute of Science Tokyo} \\
Tokyo, Japan \\
cao@c.titech.ac.jp}
\and
\IEEEauthorblockN{Hongzhi Yin}
\IEEEauthorblockA{\textit{The University of Queensland} \\
Brisbane, Australia \\
h.yin1@uq.edu.au}
}

\maketitle
 
\begin{abstract}
Time series forecasting plays a critical role in various real-world applications, including energy consumption prediction, disease transmission monitoring, and weather forecasting. Although substantial progress has been made in time series forecasting, most existing methods rely on a centralized training paradigm, where large amounts of data are collected from distributed devices (e.g., sensors, wearables) to a central cloud server. However, this paradigm has overloaded communication networks and raised privacy concerns. Federated learning, a popular privacy-preserving technique, enables collaborative model training across distributed data sources. However, directly applying federated learning to time series forecasting often yields suboptimal results, as time series data generated by different devices are inherently heterogeneous.
In this paper, we propose a novel framework, \modelname, to address data heterogeneity by generating informative synthetic data as auxiliary knowledge carriers. 
Specifically, \modelname generates two types of synthetic data. 
The first type of synthetic data captures the representative distribution information from clients' uploaded model updates and enhances clients' local training consensus.
The second kind of synthetic data extracts long-term influence insights from global model update trajectories and is used to refine the global model after aggregation. 
\modelname is compatible with most time series forecasting models and can be seamlessly integrated into existing federated learning frameworks to improve prediction performance. Extensive experiments on eight datasets, using several federated learning baselines and four popular time series forecasting models, demonstrate the effectiveness and generalizability of \modelname.
\end{abstract}


\section{Introduction}\label{sec_introduction}
With the proliferation of sensors, wearables, and Internet of Things (IoT) devices, the volume of time series data has increased dramatically in recent years. Time series forecasting has emerged as a focal point for both academic and industrial communities, reflecting its importance of automatically extracting meaningful patterns from extensive historical data to predict future values. Existing forecasting methods primarily aim to enhance prediction accuracy by employing advanced deep learning architectures to model temporal dependencies~\cite{li2023towards}. For example, several studies~\cite{zhou2021informer,liuitransformer} have refined the Transformer architecture~\cite{vaswani2017attention} to efficiently handle long-sequence time series predictions. Meanwhile, recent research~\cite{zhang2022less,zeng2023transformers,chentsmixer} has explored the use of MLPs for capturing temporal information, achieving state-of-the-art performance.

While the aforementioned models have achieved significant success, most rely on centralized training, where large volumes of data are collected from widely deployed devices and uploaded to a central server or cloud. However, collecting data from distributed devices presents practical challenges due to limited bandwidth and stringent privacy regulations (e.g., GDPR~\footnote{\url{https://gdpr-info.eu/}} and CCPA~\footnote{\url{https://oag.ca.gov/privacy/ccpa}}). For example, electricity usage data can reveal highly sensitive information about individuals, causing data owners to hesitate to share it due to privacy concerns~\cite{asghar2017smart}. Similarly, smart wearables record detailed personal health metrics, such as oxygen saturation, heart rate, ECG, and EEG. Given the sensitive nature of this data, sharing it significantly heightens the risk of data breaches.

Federated learning~\cite{mcmahan2017communication,li2022federated} offers a privacy-preserving framework for training predictive models without sharing or transmitting raw data. In this approach, a central server coordinates multiple clients, enabling them to collaboratively train models on their locally stored data. This ensures data privacy while minimizing the need to transmit large volumes of raw data.  In this work, we focus on cross-device federated time series forecasting, where each device operates as an individual client due to its growing computational capabilities. This federated framework provides robust privacy protection by ensuring that data remains stored locally on each device. Throughout the subsequent sections, the terms ``client'' and ``device'' are used interchangeably.
However, traditional federated learning methods assume that data across clients follow independent and identical distributions (IID), which is rarely true in time series data. Time series data are inherently heterogeneous, as they are generated by diverse devices operating under varying conditions and severing various applications. Existing federated learning methods face significant challenges in effectively learning from such heterogeneous data~\cite{ye2023heterogeneous}.

Two key scenarios give rise to heterogeneity in federated time series forecasting. (1) \textbf{Multivariate Time Series Forecasting}: This scenario occurs when different variables or dimensions of the same entities are collected and stored on separate devices. These variables are inherently heterogeneous and correlated. However, leveraging and integrating their correlations can significantly enhance forecasting performance across all variables, as evidenced by numerous multivariate time series prediction models developed using traditional centralized approaches~\cite{chentsmixer}. In federated time series forecasting, models are trained independently on each variable, with model aggregation serving as the sole mechanism for information sharing. Unfortunately, this approach fails to capture the complex relationships between variables and can even degrade forecasting accuracy for individual variables after aggregation. (2) \textbf{Heterogeneous Temporal Patterns}: This scenario arises when different devices monitor the same variables or dimensions across distinct entities. For example, electricity usage data collected by smart meters varies significantly across households due to differences in lifestyle and living activities. Aggregating models that capture heterogeneous temporal patterns or distributions often produce a suboptimal global model. Addressing these heterogeneity challenges in cross-device federated time series forecasting remains a largely unexplored research area.

To address the challenges of learning from heterogeneous time series data, we propose a novel method \modelname (\underline{Fed}erated \underline{T}ime Se\underline{r}ies For\underline{e}casting with Sy\underline{n}thetic \underline{D}ata). 
\modelname draws inspiration from recent advancements in data condensation~\cite{cazenavette2022dataset,yu2023dataset,gao2024graph}, where synthetic data is generated from model trajectories to encapsulate the essential information. 
Specifically, \modelname generates two types of synthetic data on the central server to tackle the two forms of heterogeneity respectively, enhancing the federated learning process. The first type of synthetic data, $\mathcal{D}_{ct}$, is generated based on model updates uploaded by all clients, encapsulating the distribution information of all clients. This synthetic data is then distributed back to each client to augment their local training alongside their own data. By doing so, each client can benefit from all the other clients' variable representative information, akin to training multivariable time series models in a centralized manner. The second type of synthetic data, $\mathcal{D}_{gt}$, is derived from global model trajectories, capturing the dynamic patterns in the global model parameters. This data is used in conjunction with client-uploaded model updates to refine the aggregation of the global model, mitigating the challenges posed by heterogeneous temporal patterns.

Noted that unlike other data condensation-based federated learning approaches~\cite{goetz2020federated,xiong2023feddm,liumeta,wang2024aggregation}, where synthetic data acts as the primary information carrier for collaborative learning, \modelname uses synthetic data solely as auxiliary information. The primary learning process remains rooted in sharing model parameters and updates, ensuring system performance does not overly depend on the quality of synthetic data - a factor often difficult to guarantee. Furthermore, the synthetic data construction occurs entirely on the central server, minimizing the computational burden on client devices.

To sum up, the major contributions of this work are as follows:
\begin{itemize}
    \item  To the best of our knowledge, we are the first to propose and conduct a comprehensive investigation into the data heterogeneity challenge in cross-device federated time series forecasting.
    \item  We propose a versatile federated time series forecasting component, \modelname, which addresses the challenges of learning from heterogeneous time series data by constructing two types of synthetic data derived from clients' uploaded models and the aggregated global models.
    \item To validate the effectiveness and generalizability of \modelname, we conducted extensive experiments on eight widely used time series forecasting datasets. By integrating \modelname into several mainstream federated learning frameworks, we train four widely adopted state-of-the-art time series forecasting models: DLinear~\cite{zeng2023transformers}, LightTS~\cite{zhang2022less}, TSMixer~\cite{chentsmixer}, and iTransformer~\cite{liuitransformer}. The results consistently demonstrate that \modelname significantly improves the federated forecasting performance across all time series forecasting datasets from diverse application scenarios.
\end{itemize}

The remainder of this work is organized as follows.
In Section~\ref{sec_relatedwork}, we present a literature review of related topics, followed by introducing the preliminaries of time series forecasting in Section~\ref{sec_preliminaries}.
Section~\ref{sec_methodology} describes the technical details of our \modelname.  Extensive empirical results and analysis are shown in Section~\ref{sec_experiments}.
Finally, a brief summarization of this paper is delivered in Section~\ref{sec_conclusion}.

\section{Related Work}\label{sec_relatedwork}
In this section, we briefly review the literature of four related topics: time series forecasting, federated learning with data heterogeneity, federated learning with data condensation, and federated time series forecasting.

\subsection{Time Series Forecasting}
Time series forecasting is a widely used task applicable to numerous real-world scenarios, including energy consumption prediction~\cite{wen2022robust}, pandemic spread modeling~\cite{shen2023forecasting}, weather and traffic forecasting~\cite{chen2023prompt}, and more~\cite{benidis2022deep}. Early studies in time series forecasting relied on statistical methods, such as autoregressive integrated moving average (ARIMA)~\cite{box1970distribution}, exponential smoothing~\cite{gardner1985exponential}, and structural models~\cite{harvey1990forecasting}. However, these methods require extensive expert effort to develop.

In recent years, deep learning-based approaches have become the dominant trend in time series forecasting~\cite{benidis2022deep}. These methods typically use neural network architectures, such as RNNs~\cite{sherstinsky2020fundamentals}, CNNs~\cite{wu2017introduction}, Transformers~\cite{vaswani2017attention}, and MLPs, as backbone models. For example, LSTNet~\cite{lai2018modeling} and TPR-LSTM~\cite{shih2019temporal} combine CNNs and RNNs with attention mechanisms to capture both short- and long-term dependencies in time series data. However, RNN-based methods often suffer from issues like gradient explosion or vanishing, while CNNs are limited in modeling long-term sequences due to the restricted size of their receptive fields.
Transformers have recently gained popularity in time series forecasting due to their ability to model global dependencies~\cite{wen2023transformers}.
Zhou et al.~\cite{zhou2021informer} introduced Informer, which reduces time complexity and enhances memory efficiency. Autoformer~\cite{wu2021autoformer} introduces a decomposition architecture with an auto-correlation mechanism, and Liu et al.~\cite{liuitransformer} innovatively reverse the data dimensions in the Transformer's attention and feed-forward layers, achieving improved performance.
The use of pure MLPs in time series forecasting has also become a recent trend because of their simplicity of implementation and effectiveness of performance. Zeng et al.~\cite{zeng2023transformers} challenged the effectiveness of Transformers in time series forecasting domain, proposing a MLP-based model called DLinear. LightTS~\cite{zhang2022less} incorporates two down-sampling methods, interval sampling and continuous sampling, to enhance MLP performance, while TSMixer~\cite{chentsmixer} uses mixing operations across both time and feature dimensions to efficiently capture relevant information.

However, all of these methods are implemented in a centralized manner, overlooking the practical challenges of data privacy in real-world applications.

\subsection{Federated Learning with Data Heterogeneity} 
Federated learning enables collaborative training of a global model without requiring access to clients' raw data and has been widely researched in many domains~\cite{nguyen2017argument,yuan2023federated,yuan2024hetefedrec,yuan2024hide,yuan2023manipulating}.
This learning paradigm has garnered significant attention for applications where data collection is challenging~\cite{yang2019federated,yin2020overcoming}. FedAvg~\cite{mcmahan2017communication} was the first and remains the most widely used federated learning framework.
It trains a global model by averaging the local models of participating clients. However, FedAvg's performance suffers when data across clients is heterogeneous, a common situation as clients independently collect data in diverse environments.

To address data heterogeneity in federated learning, numerous methods have been proposed~\cite{li2022federated,ye2023heterogeneous}. Broadly, these approaches can be divided into two categories based on whether they maintain compatibility with the original FedAvg protocol. For methods incompatible with FedAvg, additional assumptions or altered learning pipelines are typically required. For example, some approaches assume the central server has access to public data~\cite{collins2021exploiting,lin2020ensemble,chenfedbe} or that data transmission between clients is allowed~\cite{liu2021feddg,yoonfedmix}. These extra requirements limit their practical applicability.
Therefore, in this paper, we focus on approaches to handling data heterogeneity within the standard federated learning framework. FedProx~\cite{li2020federated} introduces a proximal term during local model training to prevent local updates from deviating too far from the global model. SCAFFOLD~\cite{karimireddy2020scaffold} employs a control variate for variance reduction to stabilize aggregation. FedDyn~\cite{acar2021federated} uses dynamic regularization to adjust each client’s objective during training. Elastic~\cite{chen2023elastic} designs an elastic aggregation approach that dampens the influence of updates to sensitive parameters. Chen et al.~\cite{chen2024fair} propose FedHEAL, incorporating a fair aggregation objective to prevent global model bias toward specific domains.

Unfortunately, most of these studies address data heterogeneity in image classification tasks. Our empirical results reveal that these approaches fail to achieve significant improvements in federated time series forecasting. This is due to the unique nature of heterogeneity in time series forecasting, which differs fundamentally from that in traditional image classification. In image classification tasks, heterogeneity typically stems from variations in label or domain distributions across clients. In contrast, heterogeneity in time series forecasting arises from differences in variable types and the complex, evolving temporal patterns of the time series data.

\subsection{Federated Learning with Data Condensation}
Data condensation aims to compress a large training dataset into a smaller, synthetic dataset~\cite{yu2023dataset} and has recently been integrated into federated learning~\cite{gao2024graph}. This integration serves two primary purposes: (1) to improve communication efficiency~\cite{zhou2020distilled,hu2022fedsynth,zhang2022dense,zheng2016keyword,daienhancing} and (2) to address data heterogeneity~\cite{goetz2020federated,xiong2023feddm,pi2023dynafed,liumeta,hung2017computing,wang2024aggregation}. Here, we focus on the latter.
Goetz et al.~\cite{goetz2020federated} and Xiong et al.~\cite{xiong2023feddm} propose a standard workflow where clients locally compress a small synthetic dataset and share it with the central server.
The server then trains a global model on the gathered synthetic data and distributes this model back to the clients. Wang et al.~\cite{wang2024aggregation} extend this approach by allowing clients to upload average logits of real data, further improving system performance. However, these methods have several limitations. First, their performance heavily relies on the quality of the synthetic data generated by each client, which is difficult to guarantee. Additionally, these approaches require clients to have substantial computational resources, as generating synthetic data is computationally intensive. In federated time series forecasting, clients are often sensors or mobile devices with limited computational capacity, making these methods less suitable for such environments. 
DynaFed~\cite{pi2023dynafed} is more closely related to our approach, \modelname, but it only uses synthetic data to adjust the global model and is designed for image classification tasks.

\subsection{Federated Learning in Time Series Forecasting}
The research of federated time series forecasting is still under-explored.
Time-FFM~\cite{liu2024time} investigates this topic at the organization level, where each data organization (e.g., traffic data organization or electrical data organization) acts as a client. Abdel et al.~\cite{abdel2024federated} apply organization-level federated learning to train a time series forecasting model based on large language models. Yan et al.\cite{yan2022multi} propose a vertical federated learning structure. 
In this paper, we propose a device-level federated time series forecasting framework that alleviates data heterogeneity by generating synthetic data.

\section{Preliminaries}\label{sec_preliminaries}
In this section, we present a formal introduction for the settings of federated time series forecasting and then briefly introduce the basic time series forecasting models.
Note that, we use squiggle uppercase (e.g., $\mathcal{A}$) to indicate set or algorithms, bold lowercase (e.g., $\mathbf{a}$) to represent vectors, and bold uppercase (e.g., $\mathbf{A}$) to denote matrices or tensors.
Table~\ref{tb_notation} lists some important notations.

\begin{table}[]
    \centering
    \caption{List of important notations.}\label{tb_notation}
    \resizebox{0.48\textwidth}{!}{
    \begin{tabular}{l|l}
    \hline
     $c_{i}$ & a client/device in federated time series forecasting \\
     $\mathcal{D}_{c_{i}}$ & the local dataset for client $c_{i}$.  \\
     $\mathcal{D}_{ct}$ & synthetic dataset generated using clients' model trajectories.  \\
     $\mathcal{D}_{gt}$ & synthetic dataset generated using global model trajectories.  \\
     $\mathcal{T}_{ct}$ & trajectories bank for client model updates.  \\
     $\mathcal{T}_{gt}$ & trajectories bank for global model updates.  \\
     \hline
     $\mathbf{X}_{j}^{c_{i}}, \mathbf{Y}_{j}^{c_{i}}$ & the $j$th input/target output data for client $c_{i}$ \\
     $\mathbf{X}^{ct}_{i}$, $\mathbf{Y}^{ct}_{i}$ & the $i$th input/target output data (trainable parameters) in $\mathcal{D}_{ct}$\\
     $\mathbf{X}^{gt}_{i}$, $\mathbf{Y}^{gt}_{i}$ & the $i$th input/target output data (trainable parameters) in $\mathcal{D}_{gt}$\\
     $\mathbf{W}_{c_{i}}^{t}$ & the model trained by $c_{i}$ at round $t$ \\
     $\mathbf{W}^{t}$ & the aggregated global model at round $t$ \\
     \hline
    $L_{x}, L_{y}$ & the input/output data length\\
    $L_{ct}$ & \makecell[l]{the update frequency of $\mathcal{D}_{ct}$ and the trajectories segment \\ length in $\mathcal{D}_{ct}$ construction} \\
    $L_{gt}$ & the update frequency of $\mathcal{D}_{gt}$ \\
    $L_{gt}^{seg}$ & the trajectories segment length in $\mathcal{D}_{gt}$ construction \\
     \hline
    \end{tabular}}
    \end{table}

\subsection{Formulation of Federated Time Series Forecasting}\label{sec_ftsf_formulation}  
Let $\mathcal{C}=\{c_{i}\}_{i=1}^{|C|}$ be the set of clients/devices and $|C|$ is the number of all clients.
For a client $c_{i}$, it owns time series data $\mathbf{X}_{c_{i}} = [\mathbf{x}^{c_{i}}_{1}, \mathbf{x}^{c_{i}}_{2}, \dots, \mathbf{x}^{c_{i}}_{T-1}, \mathbf{x}^{c_{i}}_{T}] \in \mathbb{R}^{T\times f}$, where $T$ is the total lengths of the data and $f$ is the number of dimensions.
Notebly, in federated time series forecasting, the time series data $\mathbf{X}_{c_{i}}$ are always kept on corresponding device and will not be accessed by any other participants.
To train a time series forecasting model, clients construct a dataset $\mathcal{D}_{c_{i}} = \{ (\mathbf{X}^{c_{i}}_{j}, \mathbf{Y}^{c_{i}}_{j})\}_{j=1}^{|\mathcal{D}_{c_{i}}|}$ based on $\mathbf{X}_{c_{i}}$, where $\mathbf{X}^{c_{i}}_{j} = [\mathbf{x}_{j}^{c_{i}}, \mathbf{x}_{j+1}^{c_{i}}, \dots, \mathbf{x}_{j + L_{x} -1}^{c_{i}}, \mathbf{x}_{j + L_{x}}^{c_{i}}] \in \mathbb{R}^{L_{x}\times f}$ is a fragment of time series data as the input of a forecasting model $\mathcal{F}(\cdot)$ and $\mathbf{Y}^{c_{i}}_{j} = [\mathbf{x}_{j + L_{x} + 1}^{c_{i}}, \mathbf{x}_{j + L_{x} + 2}^{c_{i}}, \dots, \mathbf{x}_{j + L_{x} + L_{y}-1}^{c_{i}}, \mathbf{x}_{j + L_{x} + L_{y}}^{c_{i}}] \in \mathbb{R}^{L_{y}\times f}$ is the target future prediction.
Then, the goal of federated time series forecasting can be described as:
\begin{equation}\label{eq_fl_obj}
    \mathop{argmin}\limits_{\mathbf{W}} \frac{1}{|\mathcal{C}|}\sum\limits_{c_{i}\in\mathcal{C}} \frac{1}{|\mathcal{D}_{c_{i}}|} \mathcal{L}(\mathbf{W}, \mathcal{D}_{c_{i}})
\end{equation}
\begin{equation}\label{eq_local_obj}
    \mathcal{L}(\mathbf{W}, \mathcal{D}) = \sum\limits_{(\mathbf{X}_{j}, \mathbf{Y}_{j}) \in \mathcal{D}} \left\|\mathbf{Y}_{j} - \mathcal{F}(\mathbf{W}, \mathbf{X}_{j})\right\|
\end{equation}
where $\mathbf{W}$ is the parameters of the forecasting model. 

Federated time series forecasting employs a central server to coordinate clients to optimize E.q.~\ref{eq_fl_obj} without accessing clients' distributed datasets $\mathcal{D}_{c_{i}}$ by transmitting and aggregating model parameters.
Specifically, clients and the central server iteratively repeat the following steps until model convergence.
At the round of $t$, a central server selects a group of clients $\mathcal{C}^{t}$ to participate in the training process and disperses a global model parameters $\mathbf{W}^{t}$ to them.
Subsequently, clients leverage the received global model parameters to initialize a local model and optimize the local model with E.q.~\ref{eq_local_obj} on their local datasets $\mathcal{D}_{c_{i}}$.
After local training, clients upload the updated model parameters $\mathbf{W}_{c_{i}}^{t}$ to the central server.
When received the updated parameters, the central server aggregates these parameters to form a new global model parameters:
\begin{equation}\label{eq_general_agg}
    \mathbf{W}^{t+1} \leftarrow agg(\{\mathbf{W}^{t}_{c_{i}}\}_{c_{i}\in \mathcal{C}^{t}})
\end{equation}
One main-stream aggregation solution is FedAvg~\cite{mcmahan2017communication}, which utilizes weighted average to aggregate client uploaded parameters:
\begin{equation}\label{eq_fedavg}
    \mathbf{W}^{t+1} =  \sum\limits_{c_{i}\in \mathcal{C}^{t}} \frac{|\mathcal{D}_{c_{i}}|}{\sum\limits_{c_{j}\in \mathcal{C}^{t}} |\mathcal{D}_{c_{j}}|}   \mathbf{W}^{t}_{c_{i}}
\end{equation}
This design performs well when the client data are homogeneous. However, when client data are heterogeneous, local models are updated towards local optimal solution and FedAvg cannot simply aggregate them to achieve optimal global performance.

\subsection{Base Time Series Forecasting Models}  
A federated time series forecasting framework should ideally be compatible with most time series forecasting models. In this paper, to demonstrate the generalizability of our proposed method, we select four recent state-of-the-art time series models that cover two major architectures: Transformer and MLP.

\textbf{DLinear}~\cite{zeng2023transformers}: DLinear decomposes input time series data into seasonal and trend components using a moving average kernel. For each component, DLinear employs a linear layer network, summing the resulting features for prediction.

\textbf{LightTS}~\cite{zhang2022less}: 
LightTS utilizes two down-sampling strategies, continuous sampling and interval sampling, to process time series data. Besides, it introduces an Information Exchange Block (IEBblock), which consists of two MLPs that encode input matrix data from both the row and column perspectives. 

\textbf{TSMixer}~\cite{chentsmixer}: 
Generally, TSMixer consists of four main components: a time-mixing MLP for capturing temporal patterns, a feature-mixing MLP for leveraging covariate information across time steps, a temporal projection layer to adjust the input length for forecasting, and residual connections that link the MLPs to enhance model depth.

\textbf{iTransformer}~\cite{liuitransformer}:
Unlike other Transformer-based forecasting models~\cite{zhou2021informer}, iTransformer retains the original Transformer architecture but inverts the input dimensions. It encodes the full history of each variable into a single token, using the attention mechanism to capture correlations between variables instead of time steps. This approach inherently encodes temporal information, making positional encoding unnecessary.

\section{Methodology}\label{sec_methodology}  
In this section, we firstly provide the overview and motivation of developing \modelname.
After that, we detailly describe the techniques of each component in \modelname.

\begin{figure*}[t]
    \centering
    \includegraphics[width=1.3\columnwidth]{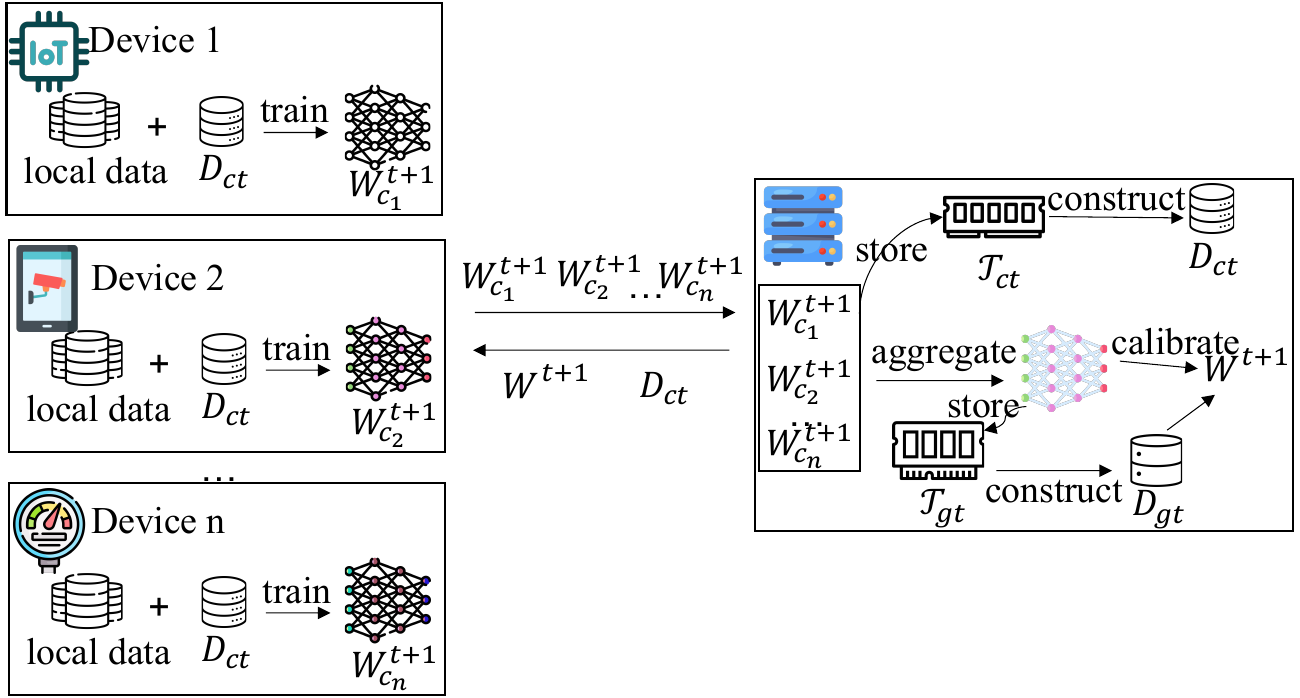} 
    \caption{The overall architecture of \modelname. When clients uploaded their model updates, these updates are (1) used for aggregation as the original federated learning and (2) stored in a trajectory bank $\mathcal{T}_{ct}$ for $\mathcal{D}_{ct}$ construction. In addition, the aggregated global model is recorded in another trajectory bank $\mathcal{T}_{gt}$, which is used to construct the synthetic data $\mathcal{D}_{gt}$. After that, $\mathcal{D}_{ct}$ is sent back to clients for local training, while $\mathcal{D}_{gt}$ is used to refine the aggregated global model.}
    \label{fig_ours}
\end{figure*}

\subsection{Overview of \modelname}
In Section~\ref{sec_introduction}, we analyze two key scenarios in federated time series forecasting: (1) the time series data on clients correspond to different variables, and (2) clients have the same variables but with distinct temporal patterns due to their unique characteristics. Based on this analysis, we identify two critical weaknesses of the original federated learning framework that hinder its ability to handle such heterogeneous scenarios.
First, clients rely solely on model aggregation for knowledge transfer, which fails to capture the complex relationships between clients, especially when these clients are corresponding to different variables. Second, the server lacks the capability to aggregate a superior global model from the uploaded client models, particularly when faced with heterogeneous data distributions.

In light of this, we introduce \modelname, a framework designed to address heterogeneity in federated time series forecasting by improving knowledge transfer among clients while calibrating a better aggregated global model.
\modelname achieves these objectives by constructing two types of synthetic data using model trajectories from various sources.
The first type of synthetic data, $\mathcal{D}_{ct}$, encapsulates the representative distribution information derived from clients' uploaded model updates. This data acts as a ``proxy" that enhances knowledge transfer by integrating with clients' local training.
The second type of synthetic data, $\mathcal{D}_{gt}$, captures the long-term dynamic changes in the aggregated global model trajectories and is used to refine the global model. An overview of \modelname is depicted in Figure~\ref{fig_ours}, and its workflow is presented in pseudocode in Algorithm~\ref{alg_fedtrend}.

Notably, our \modelname is compatible with most federated learning frameworks as it addresses heterogeneity from the synthetic data construction aspect and does not break the basic federated learning protocol.
Without loss of generality, we introduce \modelname based on the most general learning framework illustrated in Section~\ref{sec_ftsf_formulation}.
In the experimental section part, we will also evaluate the empirical performance of \modelname when integrated with various federated learning frameworks.

\subsection{Synthetic Data $\mathcal{D}_{ct}$ for Representative Knowledge Transfer}
The recently developed technique of data condensation~\cite{cazenavette2022dataset} has demonstrated that a synthetic dataset can be learned from model training trajectories to summarize useful information.
Building on this idea, \modelname introduces a synthetic dataset, $\mathcal{D}_{ct}$, designed to capture representative knowledge of all clients from clients' uploaded model parameters.

Specifically, at each federated training round $t$, when client $c_{i}$ uploads its updated model parameters $\mathbf{W}_{c_{i}}^{t}$, except for using this model updates for model aggregation, the central server will also store them in a trajectories bank $\mathcal{T}_{ct} = \{c_{i}: [\mathbf{W}_{c_{i}}^{1}, \dots, \mathbf{W}_{c_{i}}^{t}] \}_{c_{i}\in \mathcal{C}}$.
Then, the central server will optimize a synthetic dataset $\mathcal{D}_{ct}$ based on $\mathcal{T}_{ct}$ as follows:
\begin{gather}
    \mathop{argmin}\limits_{\mathcal{D}_{ct}} \mathop{\mathbb{E}}\nolimits_{c_{j}\sim U(\mathcal{C}), s\sim U(1, t-L_{ct})} \left[ d(\mathbf{\widetilde{W}}_{c_{j}}^{s+L_{ct}}, \mathbf{W}_{c_{j}}^{s+L_{ct}})  \right]\label{eq_ct_outter} \\
    s.t. \mathbf{\widetilde{W}}_{c_{j}}^{s+L_{ct}} = \mathop{argmin} \mathcal{L}(\mathbf{W}_{c_{j}}^{s}, \mathcal{D}_{ct})\label{eq_ct_inner}
\end{gather}
Here, $L_{ct}$ is the segment length of a trajectory and $U(\cdot)$ denotes uniform random sampling.
The meaning of the above two formulas is that, in E.q.~\ref{eq_ct_inner}, we train a forecasting model initialized from $\mathbf{W}_{c_{j}}^{s}$ on $\mathcal{D}_{ct} = \{(\mathbf{X}^{ct}_{i}, \mathbf{Y}^{ct}_{i})\}_{i=1}^{|\mathcal{D}_{ct|}}$ for $L_{c}$ steps, obtaining the trained model $\mathbf{\widetilde{W}}_{c_{j}}^{s+L_{ct}}$.
Note that same as the settings in real dataset, $\mathbf{X}^{ct}_{i}$ and $\mathbf{Y}^{ct}_{i}$ are the input and target output pair.
Then, in E.q.~\ref{eq_ct_outter}, we minimize the distance between $\mathbf{\widetilde{W}}_{c_{j}}^{s+L_{ct}}$ and $\mathbf{W}_{c_{j}}^{s+L_{ct}}$ via optimizing $\mathcal{D}_{ct}$, i.e., the synthetic time series data pair $(\mathbf{X}^{ct}_{i}, \mathbf{Y}^{ct}_{i})$ are learnable parameters.

Intuitively, by optimizing Eqs.~\ref{eq_ct_outter} and~\ref{eq_ct_inner}, we can obtain a synthetic dataset $\mathcal{D}_{ct}$, where initializing a model with any client's model updates from $\mathcal{T}{ct}$ and subsequently training the model on $\mathcal{D}_{ct}$ produces updates similar to those obtained by training on the clients' original local data. 
In other words, $\mathcal{D}_{ct}$ effectively captures the essential information of all clients' local data for training their local models from initialization to round $t$.

Ideally, the optimization should be performed every time when clients upload new model updates. However, constructing the synthetic dataset is computationally intensive due to its bi-level optimization process, posing a heavy burden on the central server. 
Since the goal of \modelname to construct $\mathcal{D}_{ct}$ is to learn the representative information among clients rather than replacing the original dataset in each client as the traditional data condensation goal, which requires a very high quality of synthetic data, we simplify the process to reduce computational and memory costs.
In detail, we only update the synthetic datasets at intervals of $L_{ct}$.
For each client $c_{i}$, the central server temporarily stores only the start and end model updates within these intervals $\mathbf{W}_{c_{j}}^{k*L_{ct}}$ and $\mathbf{W}_{c_{j}}^{(k+1)*L_{ct}}$, i.e., $\mathcal{T}_{ct} = \{c_{i}: (\mathbf{W}_{c_{j}}^{k*L_{ct}}, \mathbf{W}_{c_{j}}^{(k+1)*L_{ct}}) \}_{c_{i}\in \mathcal{C}}$.
Then, when the round $t = (k+1)*L_{ct}$, the central server will construct a synthetic dataset $\mathcal{D}_{ct}^{t}$ based on the trajectories $\mathcal{T}_{ct}$.
That is to say, E.q.~\ref{eq_ct_outter} is simplied to:
\begin{equation}\label{eq_ct_outter_efficiency}
    \mathop{argmin}\limits_{\mathcal{D}_{ct}} \mathop{\mathbb{E}}\nolimits_{c_{j}\sim U(\mathcal{C})} \left[ d(\mathbf{\widetilde{W}}_{c_{j}}^{k*L_{ct}}, \mathbf{W}_{c_{j}}^{(k+1)*L_{ct}})  \right]
\end{equation}
After obtaining the optimized $\mathcal{D}_{ct}$, $\mathcal{T}_{ct}$ is set to empty and used to record the next pair of start and end parameters of $L_{ct}$ length model trajectories.
Therefore, it can also reduce the memory burden as we only need the store a pair of model updates for clients, rather than clients' whole model updates.

Moreover, to further reduce the optimization difficulty and make the synthetic dataset focus on learning significant information from trajectories $\mathcal{T}_{ct}$, we only utilize the model parameters that consistently update towards the same directions as the learning sources for $\mathcal{D}_{ct}^{t}$ optimization.
According to~\cite{chen2024fair}, the parameters that consistently update towards a direction may reflect some important signals and learn from these parameters can make the synthetic dataset more concentrate on extracting these knowledge.
Therefore, before the parameters $\mathbf{\widetilde{W}}_{c_{j}}^{k*L_{ct}}$ been added to the trajectories memory bank $\mathcal{T}_{ct}$, the central server will firstly check whether the gradient of the element $w_{c_{j}, m} \in \mathbf{\widetilde{W}}_{c_{j}}^{k*L_{ct}}$ is consistent with its previous updates, i.e., $sign(\Delta w_{c_{j}, m}^{k*L_{ct} -1}) == sign(\Delta w_{c_{j}, m}^{k*L_{ct}})$.
If the update is not consistent, the distance loss of corresponding element in E.q.~\ref{eq_ct_outter_efficiency} will be masked.

After the central server constructed the synthetic dataset $\mathcal{D}_{ct}$, it will be dispersed to each client and mixed with clients' local dataset for local training.
Since the dataset $\mathcal{D}_{ct}$ is optimized on the consistent model trajectories from all clients, it captures the representative knowledge of all clients' local data. 
Thus, incorporating this synthetic dataset helps mitigate local data heterogeneity, enabling clients to learn from each other indirectly.

\begin{algorithm}[!ht]
    \renewcommand{\algorithmicrequire}{\textbf{Input:}}
    \renewcommand{\algorithmicensure}{\textbf{Output:}}
    \caption{The pseudo-code for \modelname.} \label{alg_fedtrend}
    \begin{algorithmic}[1]
      \Require global round $R$; learning rate $lr$, $L_{ct}$, $L_{gt}$ \dots
      \Ensure  well-trained time series forecasting model $\mathbf{W}^{R}$
      \State server initializes model $\mathbf{W}^{0}$
      \State $\mathcal{T}_{ct} = \{c_{i}:[\mathbf{W}^{0}_{c_{i}}]\}_{c_{i}\in \mathcal{C}}$, $\mathcal{T}_{gt} = \{\mathbf{W}^{0}\}$
      \State $\mathcal{D}_{ct}=\emptyset, \mathcal{D}_{gt}=\emptyset$
      \For {each round $t = 0, ..., R-1$}
        \State sample a fraction of clients $\mathcal{C}^{t}$ from $\mathcal{C}$
          \For{$c_{i}\in \mathcal{C}^{t}$ \textbf{in parallel}} 
          \State // execute on client sides
          \State $\mathbf{W}_{c_{i}}^{t+1}\leftarrow$\Call{ClientTrain}{$c_{i}$, $\mathbf{C}^{t}$, $\mathcal{D}_{ct}$}
          \EndFor
        \State // execute on central server
        \If{$t \% L_{ct} == 0$}
            \State check each element's update direction consistency
            \State append $\mathbf{W}_{c_{i}}^{t+1}$ into $\mathcal{T}_{ct}$
          \EndIf
        \State $\mathbf{W}^{t+1}\leftarrow$ aggregate received client model parameters $\{\mathbf{W}_{c_{i}}^{t+1}\}_{c_{i}\in \mathcal{C}^{t}}$
        \State append $\mathbf{W}^{t+1}$ into $\mathcal{T}_{gt}$
        \State refine $\mathbf{W}^{t+1}$ on $\mathcal{D}_{gt}$
      \EndFor
      \If{$t \% L_{ct} == 0$}
            \State $\mathcal{D}_{ct}\leftarrow$ \Call{DataConstruction}{$\mathcal{T}_{ct}$}
      \EndIf 
      \If{$t \% L_{gt} == 0$}
            \State $\mathcal{D}_{gt}\leftarrow$ \Call{DataConstruction}{$\mathcal{T}_{gt}$}
      \EndIf
      \Function{ClientTrain} {$c_{i}$, $\mathbf{W}^{t}$, $\mathcal{D}_{ct}$}
      \State download $\mathbf{W}^{t}$ and $\mathcal{D}_{ct}$
      \State $\mathbf{W}_{c_i}^{t+1}\leftarrow$ update local model with forecasting objective E.q.~\ref{eq_local_obj} on $\mathcal{D}_{c_{i}}$ and $\mathcal{D}_{ct}$
      \State \Return $\mathbf{W}_{c_{i}}^{t+1}$
      \EndFunction
      \Function{DataConstruction} {$\mathcal{T}$}
      \State randomly initialize synthetic dataset $\mathcal{D}_{syn}$
      \For {synthetic data training iteration $n = 1\dots N$}
        \State sample a segment of trajectories $(\mathbf{W}^{start}, \mathbf{W}^{end})$ from trajectories bank $\mathcal{T}$
        \State $\widetilde{\mathbf{W}}^{end} \leftarrow$ train $\mathbf{W}^{start}$ on $\mathcal{D}_{syn}$
        \State compute distance loss $d(\widetilde{\mathbf{W}}^{end}, \mathbf{W}^{end})$ and gradient based on $\mathcal{D}_{syn}$
      \EndFor
      \State \Return $\mathcal{D}_{syn}$
      \EndFunction

      \end{algorithmic}
  \end{algorithm}

\subsection{Synthetic Data for Global Model Refinement}
Although the synthetic data $\mathcal{D}_{ct}$ can carry some representative information from all clients to improve the local training consensus, the heterogeneity problem still persists.
This is because that the size of $\mathcal{D}_{ct}$ is limited considering the communication cost and it is hard to let $\mathcal{D}_{ct}$ capture all information of clients' data.
Consequently, the global model may still drift away from the optimal point after aggregation. 
To address this, \modelname introduces an additional synthetic dataset $\mathcal{D}_{gt}$ to refine the aggregated global model.

Specifically, \modelname constructs $\mathcal{D}_{gt}$ based on the trajectories of aggregated global models, so that the dataset can capture the long-term dynamics of mutual influences of clients' models trained on their heterogeneous local data.
Formally, the central server maintains a trajectory bank $\mathcal{T}_{gt} = \{\mathbf{W}^{1},\dots, \mathbf{W}^{t}\}$, storing aggregated global model $\mathbf{W}^{t}$ at each round.
Then, $\mathcal{D}_{gt}$ is optimized as follows:
\begin{gather}
    \mathop{argmin}\limits_{\mathcal{D}_{gt}} \mathop{\mathbb{E}}\nolimits_{s\sim U(1, t-L_{gt}^{seg})} \left[ d(\mathbf{\widetilde{W}}^{s+L_{gt}^{seg}}, \mathbf{W}^{s+L_{gt}^{seg}})  \right]\label{eq_gt_outter} \\
    s.t. \mathbf{\widetilde{W}}^{s+L_{gt}^{seg}} = \mathop{argmin} \mathcal{L}(\mathbf{W}^{s}, \mathcal{D}_{gt})\label{eq_gt_inner}
\end{gather}
where $L_{gt}^{seg}$ is the length of the trajectory segment and $\mathbf{\widetilde{W}}^{s+L_{gt}^{seg}}$ is the model trained on $\mathcal{D}_{gt}$ from the initialization point of $\widetilde{W}^{s}$ for $L_{gt}^{seg}$ steps.
Similar to $\mathcal{D}_{ct}$, $\mathcal{D}_{gt}$ is constructed with pairs of learnable input and target outputs $\mathcal{D}_{gt} = \{(\mathbf{X}^{gt}_{i}, \mathbf{Y}^{gt}_{i})\}_{i=1}^{|\mathcal{D}_{gt|}}$  and has been optimized using E.q.~\ref{eq_gt_outter}.
For computational efficiency, $\mathcal{D}_{gt}$ is updated at intervals of $L_{gt}$ rounds, similar to the update strategy for $\mathcal{D}_{ct}$.

Once constructed, $\mathcal{D}_{gt}$ effectively summarizes the stable influence of clients' data distributions on model aggregation within the recent $L_{gt}$ federated learning rounds.
Then, for the future aggregated global model $\mathbf{W}^{t}$, we refine it by finetuning on $\mathcal{D}_{gt}$ for calibration.

\subsection{Implementation of Synthetic Data Construction}
We employ the most commonly used synthetic data construction algorithm MTT~\cite{cazenavette2022dataset} to construct both $\mathcal{D}_{ct}$ and $\mathcal{D}_{gt}$, considering its state-of-the-art performance for meaningful data construction.
Note that \modelname is also compatible with other data construction algorithms, such as distribution DC~\cite{zhao2020dataset}, PP~\cite{li2024dataset}, FTD~\cite{du2023minimizing}, and so on.
The detailed steps for synthetic data construction are outlined in Algorithm~\ref{alg_fedtrend}, Lines 29-36. 
The integration of $\mathcal{D}_{ct}$ and $\mathcal{D}_{gt}$ into the general federated learning framework is described in Lines 11-23 and Line 26 in Algorithm~\ref{alg_fedtrend}.

\subsection{Discussion} 
In this part, we briefly discuss \modelname from three aspects: privacy, communication and computational burden.

\subsubsection{Privacy Analysis}
The synthetic data in \modelname are constructed in accordance with the standard federated learning protocol without any additional assumptions. Hence, the privacy-preserving capabilities of \modelname should be consistent with those of traditional federated learning methods. In Section~\ref{sec_rq5_privacy}, we also showcase the compatibility of \modelname with existing privacy-preserving mechanisms~\cite{yin2021comprehensive}.

\subsubsection{Communication Cost Analysis}
\modelname introduces some additional communication overhead because the central server needs to distribute $\mathcal{D}_{ct}$ to clients. However, this cost is minimal. Taking our experimental setup as an example, the central server sends $\mathcal{D}_{ct}$ to clients at intervals of $10$ rounds, with $\mathcal{D}_{ct}$ consisting of $20$ input-output pairs. Therefore, for the entire training process (spanning $80$ rounds), the extra communication cost per client is $8 \times 20 \times (size(\mathbf{X}^{ct}) + size(\mathbf{Y}^{ct}))$. Given that $size(\mathbf{X}^{ct})$ is a sequence of $24$ numbers in our case, the total additional cost is less than $30$KB per client.

\subsubsection{Computational Burden Analysis}
Existing condensation techniques often suffer from high computational costs. To mitigate this issue in federated learning systems, unlike other works~\cite{goetz2020federated,xiong2023feddm,liumeta,wang2024aggregation} that require clients to perform synthetic data construction, \modelname offloads the entire synthetic data construction process to the central server. This design choice is based on the fact that in practical applications, clients typically have limited computational resources, whereas the central server usually possesses ample computational power. As a result, the computational burden on clients in \modelname remains unchanged, while the extra load on the central server is manageable, especially considering its substantial resources and the potential for performance improvement.

\section{Experiments}\label{sec_experiments} 
In this section, we conduct experiments to answer the following research questions:
\begin{itemize}
    \item \textbf{RQ1.} How effective is our \modelname compared to existing federated learning baselines?
    \item \textbf{RQ2.} How is the generalization ability of our \modelname?
    \item \textbf{RQ3.} How does \modelname benefit from each key component?
    \item \textbf{RQ4.} How does the value of some important hyper-parameters (e.g., synthetic data sizes, frequency of synthetic data updates, input and output data lengths) affect \modelname's performance?  
    \item \textbf{RQ5.} Further study: how is the performance of \modelname with privacy protection mechanism?
\end{itemize}

\begin{table*}[htbp]
    \centering
    \caption{The statistics of datasets.} \label{tb_statistics}
    \begin{tabular}{l|ccccc}
    \hline
    \textbf{Dataset}      & \textbf{Client Num} & \textbf{Timesteps} & \textbf{Granularity} & \textbf{Domain} & \textbf{Start Time} \\ \hline
    \textbf{ETTh1, ETTh2} & 7                   & 14,400             & 1 hour               & Energy     & 2016/7/1            \\
    \textbf{Electricity}  & 321                 & 26,304             & 1 hour               & Energy     & 2012/1/1            \\
    \textbf{Traffic}      & 862                 & 17,544             & 1 hour               & Traffic         & 2015/1/1            \\
    \textbf{Solar Energy} & 137                 & 52,560              & 10 minutes           & Energy          & 2006/1/1            \\
    \textbf{State-ILI}    & 37                  & 345                & 1 week               & Disease         & 2009/10/4           \\
    \textbf{Country-Temp} & 131                 & 2,277               & 1 month              & Climate         & 1823/1/1            \\
    \textbf{USWeather}    & 12                  & 35,064              & 1 hour               & Climate         & 2010/1/1            \\ \hline
    \end{tabular}
    \end{table*}

\subsection{Datasets}
We validate \modelname on eight datasets (ETTh1, ETTh2, Electricity, Traffic, Solar Energy, State-ILI, Country-Temp, and USWeather), covering four different domains (energy, traffic, disease, and climate forecasting).
The statistics of these datasets are listed in Table~\ref{tb_statistics}.
ETTh~\footnote{\url{https://github.com/zhouhaoyi/ETDataset}} datasets record hourly electrical transformer statistics over a span of two years. 
Electricity~\footnote{\url{https://archive.ics.uci.edu/ml/datasets/ElectricityLoadDiagrams20112014}} captures the electricity consumption in kilowatts of $321$ clients from 2012 to 2014.
Traffic~\footnote{\url{http://pems.dot.ca.gov}} tracks the hourly road occupancy rate using $862$ sensors across the San Francisco Bay Area freeways over 48 months (2015-2016).
Solar Energy~\footnote{\url{http://www.nrel.gov/grid/solar-power-data.html}} provides solar power production records at 10-minute intervals in 2006 from $137$ PV plants in Alabama State.
State-ILI~\footnote{\url{https://github.com/emilylaiken/ml-flu-prediction}} tracks Influenza-like Illness (ILI) data weekly for $37$ U.S. states from approximately 2009 to 2017. 
Country-Temp~\footnote{\url{https://data.world/data-society/global-climate-change-data}} contains the average land temperature of $131$ countries from 1823 to 2013. 
USWeather~\footnote{\url{https://www.ncei.noaa.gov/data/local-climatological-data/}} includes 4 years from 2010 to 2013 climatological data and we follow the usage in~\cite{zhou2021informer}.
For each dataset, 70\% of the data is used for training and validation, while the remaining 30\% is reserved for testing. 
In this paper, we focus on cross-device federated time series forecasting, where each device (e.g., sensors, meters, wearables, etc) is treated as a client. 
Consequently, in ETTh1, ETTh2, and USWeather, clients track different variables, whereas in the remaining datasets, clients record the same variable for different entities.

\subsection{Evaluation Metrics}
Following previous time series forecasting studies~\cite{zeng2023transformers,liuitransformer,zhang2022less,chentsmixer}, we use Mean Squared Error (MSE) and Mean Absolute Error (MAE) as the primary metrics to evaluate model performance. MSE measures the average of the squared differences between the predicted and actual values, while MAE calculates the average of the absolute differences between the forecasted and actual values. Lower MSE and MAE values indicate better forecasting accuracy.

\begin{table*}[htbp]  
    \centering
    \caption{The comparison of the overall performance of \modelname and baselines. The best values of federated learning methods on each dataset are bold.}\label{tb_rq1}
    \begin{tabular}{lcccccccc}
    \hline
    \textbf{Dataset} & \multicolumn{2}{c}{\textbf{Electricity}}                            & \multicolumn{2}{c}{\textbf{Traffic}}                                & \multicolumn{2}{c}{\textbf{Solar Energy}}                           & \multicolumn{2}{c}{\textbf{State-ILI}}                              \\ \hline
    \textbf{Metrics} & \multicolumn{1}{c}{\textbf{MSE}} & \multicolumn{1}{c}{\textbf{MAE}} & \multicolumn{1}{c}{\textbf{MSE}} & \multicolumn{1}{c}{\textbf{MAE}} & \multicolumn{1}{c}{\textbf{MSE}} & \multicolumn{1}{c}{\textbf{MAE}} & \multicolumn{1}{c}{\textbf{MSE}} & \multicolumn{1}{c}{\textbf{MAE}} \\ \hline
    \textbf{Centralized}      & 0.21822                          & 0.30709                          & 0.47615                          & 0.40192                          & 0.32125                          & 0.43077                          & 0.89061                          & 0.68811                          \\
    \textbf{FedAvg}           & 0.22199                          & 0.31156                          & 0.48622                          & 0.41120                          & 0.32251                          & 0.43210                          & 0.96516                          & 0.72040                          \\
    \textbf{FedProx}          & 0.22288                          & 0.31273                          & 0.48815                          & 0.41281                          & 0.32588                          & 0.43608                          & 0.96614                          & 0.72401                          \\
    \textbf{FedDyn}           & 0.21991                          & 0.31628                          & 0.48023                          & 0.40865                          & 0.33385                          & 0.44509                          & 0.96454                          & 0.72014                          \\
    \textbf{Elastic}          & 0.21450                          & 0.30289                          & 0.47445                          & 0.39996                          & 0.32352                          & 0.43345                          & 0.96387                          & 0.71986                          \\
    \textbf{FedHEAL}          & 0.22061                          & 0.31261                          & 0.48187                          & 0.40621                          & 0.32237                          & 0.43123                          & 0.96489                          & 0.72027                          \\
    \textbf{DynaFed}          & 0.23515                          & 0.33116                          & 0.53152                          & 0.44961                          & 0.32481                          & 0.43468                          & 0.95894                          & 0.71776                          \\
    \textbf{Ours}    & \textbf{0.20888}                 & \textbf{0.29838}                 & \textbf{0.46073}                 & \textbf{0.38698}                 & \textbf{0.31457}                 & \textbf{0.42284}                 & \textbf{0.91795}                 & \textbf{0.70034}                 \\ \hline
    \textbf{Dataset} & \multicolumn{2}{c}{\textbf{Country-Temp}}                           & \multicolumn{2}{c}{\textbf{ETTh1}}                                  & \multicolumn{2}{c}{\textbf{ETTh2}}                                  & \multicolumn{2}{c}{\textbf{USWeather}}                              \\ \hline
    \textbf{Metrics} & \multicolumn{1}{c}{\textbf{MSE}} & \multicolumn{1}{c}{\textbf{MAE}} & \multicolumn{1}{c}{\textbf{MSE}} & \multicolumn{1}{c}{\textbf{MAE}} & \multicolumn{1}{c}{\textbf{MSE}} & \multicolumn{1}{c}{\textbf{MAE}} & \multicolumn{1}{c}{\textbf{MSE}} & \multicolumn{1}{c}{\textbf{MAE}} \\ \hline
    \textbf{Centralized}      & 0.26942                          & 0.37543                          & 0.37308                          & 0.40949                          & 0.15794                          & 0.27620                          & 0.45217                          & 0.44744                                \\
    \textbf{FedAvg}           & 0.54606                          & 0.57863                          & 0.39343                          & 0.42228                          & 0.16318                          & 0.28154                          & 0.45444                          & 0.45011                          \\
    \textbf{FedProx}          & 0.54606                          & 0.57983                          & 0.39646                          & 0.42428                          & 0.16524                          & 0.28367                          & 0.45533                          & 0.45123                          \\
    \textbf{FedDyn}           & 0.54044                          & 0.57762                          & 0.38909                          & 0.41953                          & 0.15655                          & 0.27625                          & 0.44625                          & 0.45352                          \\
    \textbf{Elastic}          & 0.63603                          & 0.63226                          & 0.38215                          & 0.41568                          & 0.16165                          & 0.28026                          & 0.45054                          & 0.44734                          \\
    \textbf{FedHEAL}          & 0.53606                          & 0.57324                          & 0.37465                          & 0.41158                          & 0.16188                          & 0.28121                          & 0.45202                          & 0.45350                          \\
    \textbf{DynaFed}          & 0.57391                          & 0.61212                          & 0.41513                          & 0.43578                          & 0.15899                          & 0.27764                          & 0.45613                          & 0.45683                          \\
    \textbf{Ours}    & \textbf{0.45429}                 & \textbf{0.54099}                 & \textbf{0.35814}                 & \textbf{0.39937}                 & \textbf{0.14449}                 & \textbf{0.26381}                 & \textbf{0.44036}                 & \textbf{0.44247}                 \\ \hline
    \end{tabular}
    \end{table*}
    
    \begin{figure*}[htbp]
        \centering
        \includegraphics[width=2.\columnwidth]{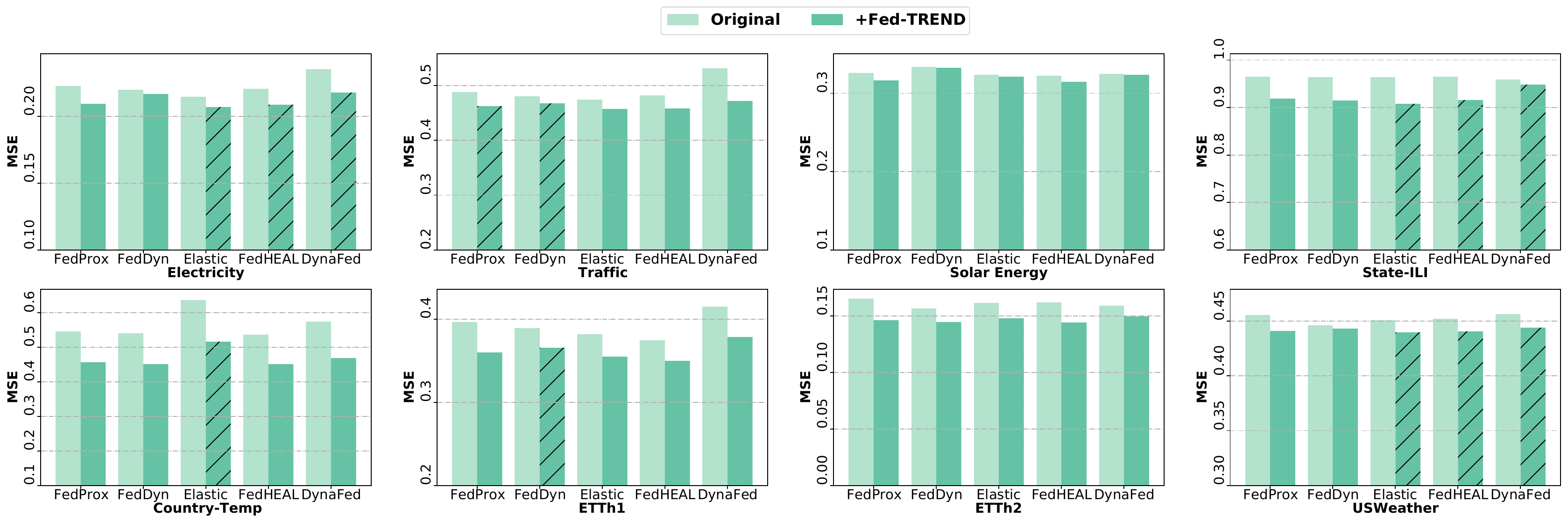} 
        \caption{The result of using \modelname to improve the general federated learning frameworks.}
        \label{fig_rq2}
    \end{figure*}

\subsection{Baselines}
To demonstrate the effectiveness of \modelname, we compare it against several baselines, including traditional centralized (Centralized), basic federated learning method (FedAvg), and state-of-the-art general federated learning solutions for data heterogeneity (FedProx, FedDyn, Elastic, FedHEAL, and DynaFed).
The following is a brief introduction for these baselines:
\begin{itemize}
    \item \textbf{Centralized}: This traditional approach trains a time series forecasting model by collecting data from all devices and training it on a central server.
    \item \textbf{FedAvg}~\cite{mcmahan2017communication}: The most widely used federated learning framework, FedAvg averages the model parameters uploaded by clients to update the global model.
    \item \textbf{FedProx}~\cite{li2020federated}: An extension of FedAvg, FedProx introduces a proximal term in the local training objective to stabilize updates from clients with diverse data distributions.
    \item \textbf{FedDyn}~\cite{acar2021federated}: FedDyn addresses data heterogeneity by adding a dynamic regularization term to the local objective function, helping synchronize client updates with the overall federated learning process.
    \item \textbf{Elastic}~\cite{chen2023elastic}: This approach handles data heterogeneity by selectively weighting or attenuating client updates, ensuring the global model benefits from stable, consistent patterns while minimizing the impact of divergent updates due to local data variations.
    \item \textbf{FedHEAL}~\cite{chen2024fair}: A state-of-the-art technique designed to address domain skew, FedHEAL maintains both local consistency and domain diversity, enhancing the global model's generalization across different client domains.
    \item \textbf{DynaFed}~\cite{pi2023dynafed}: Similar to our $\mathcal{D}_{gc}$ construction process, DynaFed is designed for image classification and does not update the synthetic data during training. Consequently, its performance degrades as the synthetic data becomes outdated over time.
\end{itemize}

\subsection{Implementation Details}
For the main experiments, we use DLinear~\cite{zeng2023transformers} as the default base model for time series forecasting in both \modelname and the baselines, considering its effectiveness and efficiency. 
In Section~\ref{sec_rq2}, we further evaluate the performance of other forecasting models within \modelname.
We set both the input and output lengths (i.e., $L_{x}$ and $L_{y}$) to $24$ and will examine the impact of these lengths in Section~\ref{sec_exp_length}. 
The total number of global training rounds $R$ is set to 80, with a local epoch count of $1$, and all clients participate in each training round. 
For local training, we use SGD~\cite{sutskever2013importance} as the optimizer, with a learning rate of $5 \times 10^{-4}$ and momentum of $0.9$. The batch size for local training is $256$.
For synthetic data generation, we set the update intervals for $\mathcal{L}_{gt}$ and $\mathcal{L}_{ct}$ to $10$, meaning the synthetic datasets $\mathcal{D}_{gt}$ and $\mathcal{D}_{ct}$ are updated every ten global rounds.
The influence of these intervals are explored in Section~\ref{sec_exp_frequency}.
The sizes of $\mathcal{D}_{gt}$ and $\mathcal{D}_{ct}$ are explored in Section~\ref{sec_exp_dct} and Section~\ref{sec_exp_dgt}, respecatively.
The optimizer for synthetic data construction is Adam~\cite{kingma2014adam} with a learning rate of $3 \times 10^{-4}$, and the number of learning iterations is set to $300$, following~\cite{cazenavette2022dataset}.

\subsection{\modelname v.s. Baselines (RQ1)}\label{sec_rq1} 
To demonstrate the effectiveness of \modelname, we compare it with several baselines in Table~\ref{tb_rq1}. The results show that the traditional federated learning framework (FedAvg) often lags behind centralized training. The performance gap varies across datasets due to differing degrees of data heterogeneity.
For example, on Solar Energy dataset, the performance between Centralized and FedAvg is very close.
This is because the data in this dataset are almost homogeneous since at any plants, the solar energy is zero at night and variations among different areas in a state are minimal.
In contrast, on datasets like State-ILI and Country-Temp datasets, intuitively, the data are highly heterogeneous, as illness statistics differ significantly across states, and ground temperatures vary widely among countries.
Consequently, the naive FedAvg approach performs much worse than centralized training in these cases.
Furthermore, most existing solutions designed to address heterogeneity in image classification do not perform well in federated time series forecasting. As discussed in the previous section, this is due to fundamental differences in task settings and the nature of data heterogeneity between image classification and time series forecasting.
Finally, our proposed method, \modelname, consistently outperforms federated learning baselines with a large margin.
Notably, on six datasets (Electricity, Traffic, Solar Energy, ETTh1, ETTh2, and USWeather), \modelname even surpasses the performance of centralized training. This improvement may be attributed to the synthetic datasets, which provide additional insights, helping the model better capture temporal patterns. On the more heterogeneous datasets, State-ILI and Country-Temp, while \modelname still has a performance gap compared to Centralized, it achieves the best results among all federated learning baselines.

\subsection{The Generalization of \modelname (RQ2)}\label{sec_rq2} 
Beyond its effectiveness, \modelname also offers strong generalization capabilities. In this section, we explore this generalizability from two perspectives:
(1) Can \modelname enhance existing federated learning frameworks by integrating with them?
and (2) Can \modelname effectively work with various time series forecasting models?

First, we investigate whether \modelname can improve the performance of federated learning baselines. 
Specifically, we integrate the construction process of $\mathcal{D}_{gt}$ and $\mathcal{D}_{ct}$ into each federated learning method's central server. 
After aggregation, $\mathcal{D}_{gt}$ is used to finetune the global model, while $\mathcal{D}_{ct}$ is mixed with local data for local training. 
Figure~\ref{fig_rq2} shows the performance results of equipping federated baselines with \modelname. 
According to the results, \modelname enhances the performance of all baselines across all datasets, demonstrating its strong generalization ability within different federated learning frameworks.

\begin{table*}[htbp]
    \centering
    \caption{The performance of \modelname with various state-of-the-art time series forecasting models.}\label{tb_model}
    \begin{tabular}{llcccccccc}
    \hline
    \textbf{}                       & \textbf{Dataset}       & \multicolumn{2}{c}{\textbf{Electricity}} & \multicolumn{2}{c}{\textbf{Traffic}}    & \multicolumn{2}{c}{\textbf{Solar Energy}}      & \multicolumn{2}{c}{\textbf{State-ILI}}    \\ \hline
    \textbf{}                       & \textbf{Metrics}       & \textbf{MSE}        & \textbf{MAE}       & \textbf{MSE}       & \textbf{MAE}       & \textbf{MSE}       & \textbf{MAE}       & \textbf{MSE}       & \textbf{MAE}       \\ \hline
    \multirow{2}{*}{\textbf{DLinear}}      & \textbf{FedAvg} & 0.22199                          & 0.31156                          & 0.48622                          & 0.41120                          & 0.32251                          & 0.43210                          & 0.96516                          & 0.72040                          \\
                                           & \textbf{Ours}   & \textbf{0.20888}                 & \textbf{0.29838}                 & \textbf{0.46073}                 & \textbf{0.38698}                 & \textbf{0.31457}                 & \textbf{0.42284}                 & \textbf{0.91795}                 & \textbf{0.70034}                 \\ \hline
    \multirow{2}{*}{\textbf{LightTS}}      & \textbf{FedAvg} & 0.22205           & 0.30894          & 0.48514          & 0.40768          & 0.33548 & 0.44212 & 1.24200          & 0.85050          \\
                                           & \textbf{Ours}   & \textbf{0.21006}   & \textbf{0.29762} & \textbf{0.46243} & \textbf{0.38758} & \textbf{0.33318}          & \textbf{0.44018}           & \textbf{1.17520} & \textbf{0.82470} \\ \hline
    \multirow{2}{*}{\textbf{TSMixer}}      & \textbf{FedAvg} & 0.21727           & 0.30388          & 0.47401          & 0.39677          & 0.27527 & 0.35847  & 1.19635          & 0.83488          \\
                                           & \textbf{Ours}   & \textbf{0.20974}  & \textbf{0.29598} & \textbf{0.46634} & \textbf{0.38831} & \textbf{0.27441}          & \textbf{0.35672}          & \textbf{0.96942} & \textbf{0.73354} \\ \hline
    \multirow{2}{*}{\textbf{iTransformer}} & \textbf{FedAvg} & 0.28840           & 0.38110           & 0.58663          & 0.47946          & 0.27552          & 0.34224           & 0.94379          & 0.72348          \\
                                           & \textbf{Ours}   & \textbf{0.28443}  & \textbf{0.37762} & \textbf{0.58295} & \textbf{0.47692} & \textbf{0.27478} & \textbf{0.34109} & \textbf{0.87321}  & \textbf{0.69234} \\ \hline
    \textbf{Dataset}                       & \textbf{}       & \multicolumn{2}{c}{\textbf{Country-Temp}} & \multicolumn{2}{c}{\textbf{ETTh1}}      & \multicolumn{2}{c}{\textbf{ETTh2}}      & \multicolumn{2}{c}{\textbf{USWeather}}  \\ \hline
    \textbf{Metrics}                       & \textbf{}       & \textbf{MSE}        & \textbf{MAE}       & \textbf{MSE}       & \textbf{MAE}       & \textbf{MSE}       & \textbf{MAE}       & \textbf{MSE}       & \textbf{MAE}       \\ \hline
    \multirow{2}{*}{\textbf{DLinear}}      & \textbf{FedAvg} & 0.54606                          & 0.57863                          & 0.39343                          & 0.42228                          & 0.16318                          & 0.28154                          & 0.45444                          & 0.45011                          \\
                                           & \textbf{Ours}   & \textbf{0.45429}                 & \textbf{0.54099}                 & \textbf{0.35814}                 & \textbf{0.39937}                 & \textbf{0.14449}                 & \textbf{0.26381}                 & \textbf{0.44036}                 & \textbf{0.44247}                 \\ \hline
    \multirow{2}{*}{\textbf{LightTS}}      & \textbf{FedAvg} & 0.71192           & 0.67898          & 0.37634          & 0.41393           & 0.16354          & 0.28388          & 0.45765                  & 0.45359                  \\
                                           & \textbf{Ours}   & \textbf{0.64298}  & \textbf{0.64576} & \textbf{0.35660} & \textbf{0.39982} & \textbf{0.15203} & \textbf{0.27366} & \textbf{0.44338}         & \textbf{0.44723}         \\ \hline
    \multirow{2}{*}{\textbf{TSMixer}}      & \textbf{FedAvg} & 0.54964           & 0.58744           & 0.37824          & 0.41354          & 0.15104           & 0.26636          & 0.44910                  & 0.45031                  \\
                                           & \textbf{Ours}   & \textbf{0.51085}  & \textbf{0.57618} & \textbf{0.36611} & \textbf{0.40612}  & \textbf{0.14236} & \textbf{0.25674} & \textbf{0.44351}         & \textbf{0.44350}         \\ \hline
    \multirow{2}{*}{\textbf{iTransformer}} & \textbf{FedAvg} & 0.88088           & 0.79160          & 0.49450          & 0.48204           & 0.16896          & 0.29863          & 0.47024                  & 0.47050                  \\
                                           & \textbf{Ours}   & \textbf{0.86617}  & \textbf{0.78476} & \textbf{0.49103} & \textbf{0.47991} & \textbf{0.16789} & \textbf{0.29747} & \textbf{0.46755}         & \textbf{0.46874}         \\ \hline
    \end{tabular}
    \end{table*}

    \begin{table*}[htbp]
        \centering
        \caption{The results of ablation study. ``-cu'' means does not construct $\mathcal{D}_{ct}$ that only considering the consistent updated gradients, i.e., using all parameters for $\mathcal{D}_{ct}$ construction. ``-$\mathcal{D}_{ct}$'' and ``-$\mathcal{D}_{gt}$'' means remove the corresponding synthetic datasets.}\label{tb_ablation}
        \begin{tabular}{lcccccccc}
        \hline
        \textbf{Dataset}                   & \multicolumn{2}{c}{\textbf{Electricity}} & \multicolumn{2}{c}{\textbf{Traffic}} & \multicolumn{2}{c}{\textbf{Solar}}  & \multicolumn{2}{c}{\textbf{Iliness}}   \\ \hline
        \textbf{Metrics}                   & \textbf{MSE}        & \textbf{MAE}       & \textbf{MSE}      & \textbf{MAE}     & \textbf{MSE}     & \textbf{MAE}     & \textbf{MSE}       & \textbf{MAE}      \\ \hline
        \textbf{\modelname} & \textbf{0.20888}    & \textbf{0.29838}   & \textbf{0.46073}  & \textbf{0.38698} & \textbf{0.31457} & \textbf{0.42284} & \textbf{0.91795}   & \textbf{0.70034}  \\
        \textbf{-cu}                   & 0.21190	             & 0.30099	            & 0.46878	           & 0.39574	          & 0.31428	          & 0.42279	          & 0.91848	            & 0.70060           \\
        \textbf{-cu -$\mathcal{D}_{ct}$}                   & 0.21524             & 0.30446            & 0.47332           & 0.39979          & 0.32014          & 0.42962          & 0.94931            & 0.71375           \\
        \textbf{-$\mathcal{D}_{gt}$}                   & 0.21645             & 0.30591            & 0.46984           & 0.39826          & 0.32307          & 0.42921          & 0.95027            & 0.71408           \\
        \textbf{-cu -$\mathcal{D}_{ct}$ -$\mathcal{D}_{gt}$ (FedAvg)}            & 0.22199             & 0.31156            & 0.48622           & 0.41120          & 0.32251          & 0.43210          & 0.96516            & 0.72040           \\ \hline
        \textbf{Dataset}                   & \multicolumn{2}{c}{\textbf{Temperature}} & \multicolumn{2}{c}{\textbf{ETTh1}}   & \multicolumn{2}{c}{\textbf{ETTh2}}  & \multicolumn{2}{c}{\textbf{USWeather}} \\ \hline
        \textbf{Metrics}                   & \textbf{MSE}        & \textbf{MAE}       & \textbf{MSE}      & \textbf{MAE}     & \textbf{MSE}     & \textbf{MAE}     & \textbf{MSE}       & \textbf{MAE}      \\ \hline
        \textbf{\modelname} & \textbf{0.45429}    & \textbf{0.54099}   & \textbf{0.35814}  & \textbf{0.39937} & \textbf{0.14449} & \textbf{0.26381} & \textbf{0.44036}   & \textbf{0.44247}  \\
        \textbf{-cu}                   & 0.45275	             & 0.53956	            & 0.37340	           & 0.41001	          & 0.15181	          & 0.27123	          & 0.44005	            & 0.44249           \\
        \textbf{-cu -$\mathcal{D}_{ct}$}                   & 0.47206             & 0.54721            & 0.38161           & 0.41520          & 0.15537          & 0.27478          & 0.44667            & 0.44655           \\
        \textbf{-$\mathcal{D}_{gt}$}                   & 0.49183             & 0.55251            & 0.36022           & 0.40132          & 0.14864          & 0.26832          & 0.44071            & 0.44281           \\
        \textbf{-cu -$\mathcal{D}_{ct}$ -$\mathcal{D}_{gt}$ (FedAvg)}            & 0.54606             & 0.57863            & 0.39343           & 0.42228          & 0.16318          & 0.28154          & 0.45444            & 0.45011           \\ \hline
        \end{tabular}
        \end{table*}

Additionally, we test \modelname's compatibility with four time series forecasting models, including two of the most popular architectures: MLP and Transformer. As shown in Table~\ref{tb_model}, \modelname improves the performance of all tested forecasting models compared to naive federated learning. Specifically, DLinear and TSMixer achieve the best performance among the four models, while iTransformer performs the worst. This is likely because, in our cross-device federated time series forecasting setting, each device tracks only a single variable. Therefore, iTransformer, which is designed to fuse information across multiple variables, is less effective in this context.

In conclusion, \modelname demonstrates strong generalizability across both federated learning frameworks and various time series forecasting models.

\begin{figure*}[htbp]
    \centering
    \includegraphics[width=2.\columnwidth]{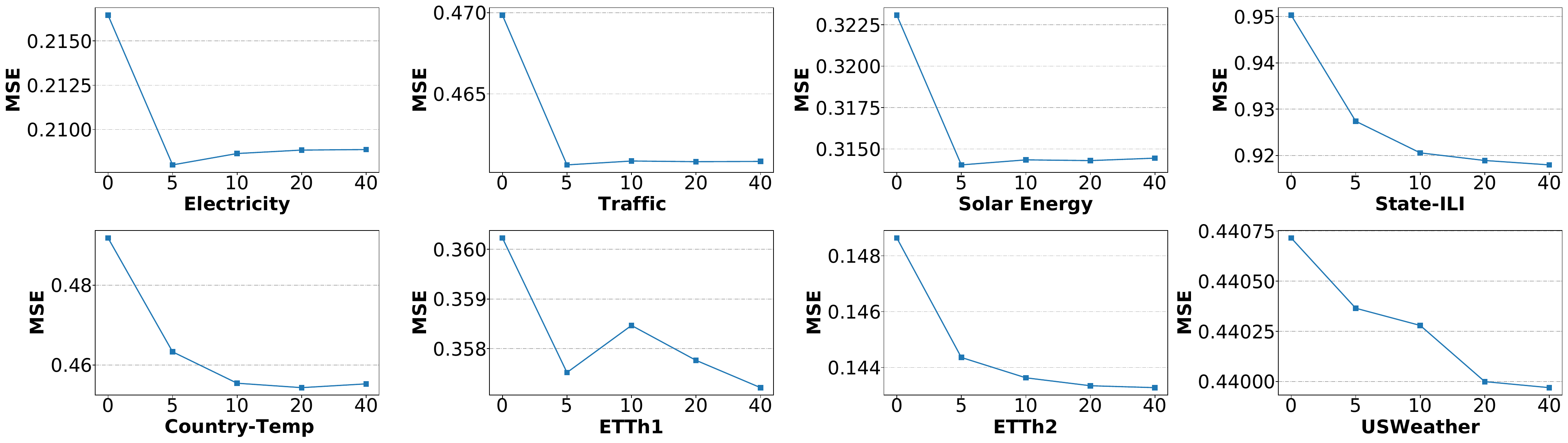} 
    \caption{The performance trend with different $|\mathcal{D}_{gt}|$.}
    \label{fig_rq5_dgt}
\end{figure*}
\begin{figure*}[htbp]
    \centering
    \includegraphics[width=2.\columnwidth]{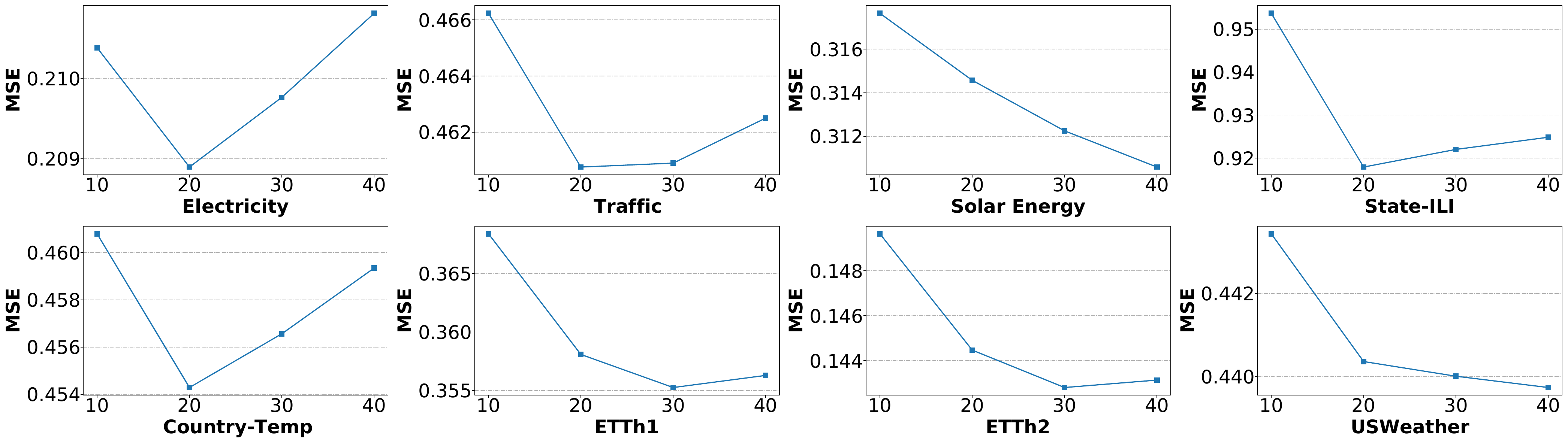} 
    \caption{The performance trend with different $|\mathcal{D}_{ct}|$.}
    \label{fig_rq5_dct}
\end{figure*}

\subsection{Ablation Studies (RQ3)}
In \modelname, we introduce the construction of two synthetic datasets: $\mathcal{D}_{ct}$, based on consistent client model updates, and $\mathcal{D}_{gt}$, based on the aggregated global model. The synthetic dataset $\mathcal{D}_{ct}$ is distributed to clients for local training, while $\mathcal{D}_{gt}$ is retained on the central server to finetune and calibrate the global model. In this section, we investigate the effectiveness of $\mathcal{D}_{gt}$, $\mathcal{D}_{ct}$, and the consistent updating (``CU'') method used in constructing $\mathcal{D}_{ct}$.

The empirical results are displayed in Table~\ref{tb_ablation}.
When we remove the consistent update dataset construction method, i.e., ``-cu'', the system's performance slightly declines. This suggests that building $\mathcal{D}_{ct}$ based on consistent updates helps the synthetic data concentrate on capturing more valuable information for model training.
Next, we examine the impact of removing the synthetic datasets $\mathcal{D}_{ct}$ (``-CU -$\mathcal{D}_{ct}$'') or $\mathcal{D}_{gt}$ (``-$\mathcal{D}_{gt}$'') individually. 
In both cases, the system's performance significantly decreases across all datasets, demonstrating the importance of each synthetic dataset. Finally, when all synthetic data components are removed, the system degrades to the FedAvg baseline.

Overall, the results indicate that each of the proposed components contributes meaningfully to improving model performance.

\subsection{Hyperparameter Analysis (RQ4)}  
In this paper, the dataset sizes $|\mathcal{D}_{gt}|$ and $|\mathcal{D}_{ct}|$ are intuitively the two most significant hyperparameters influencing the performance of \modelname. Therefore, we analyze their impact in Section~\ref{sec_exp_dgt} and Section~\ref{sec_exp_dct}, respectively. Additionally, we explore \modelname's performance with different input and output lengths in Section~\ref{sec_exp_length}, as these settings are crucial for practical time series forecasting tasks.
Last but not least, we analyze the frequency of updating $\mathcal{D}_{gt}$ and $\mathcal{D}_{ct}$, i.e., the value of $L_{gt}$ and $L_{ct}$'s influence in Section~\ref{sec_exp_frequency}.

\subsubsection{The Impact of $\mathcal{D}_{gt}$ dataset size}\label{sec_exp_dgt}  
Figure~\ref{fig_rq5_dgt} illustrates the performance trend as the size of the synthetic dataset $|\mathcal{D}_{gt}|$ increases. Across all datasets, as $|\mathcal{D}_{gt}|$ grows from $0$ to $40$, model performance improves, but the rate of improvement gradually decreases. Notably, when $|\mathcal{D}_{gt}|$ increases from $0$ to $5$, system performance improves rapidly, highlighting the positive impact of $\mathcal{D}_{gt}$. However, as $|\mathcal{D}_{gt}|$ continues to grow, the contributions to performance become minimal. This may be because, beyond a certain threshold, additional synthetic data does not provide significant new information.

\subsubsection{The Impact of $\mathcal{D}_{ct}$ dataset size}\label{sec_exp_dct}  
Figure~\ref{fig_rq5_dct} shows the performance trend of \modelname as the size of $|\mathcal{D}_{ct}|$ increases from $10$ to $40$. The results indicate that for most datasets, as the size of $\mathcal{D}_{ct}$ grows, model performance initially improves, reaching a peak. This suggests that $\mathcal{D}_{ct}$ provides valuable information for local model training. However, beyond a certain point, further expansion of the dataset size leads to a decline in performance. This phenomenon can be attributed to two main reasons. First, larger synthetic datasets introduce more trainable parameters, increasing the complexity of training and potentially capturing noise. Second, an overly large synthetic dataset may dilute the semantics of clients' original data, ultimately hindering local training. Therefore, selecting an appropriate size for $\mathcal{D}_{ct}$ is crucial for maximizing model performance.

\subsubsection{The Impact of Input and Output Time Series Data Length}\label{sec_exp_length}
\begin{figure*}[htbp]
    \centering
    \includegraphics[width=2.\columnwidth]{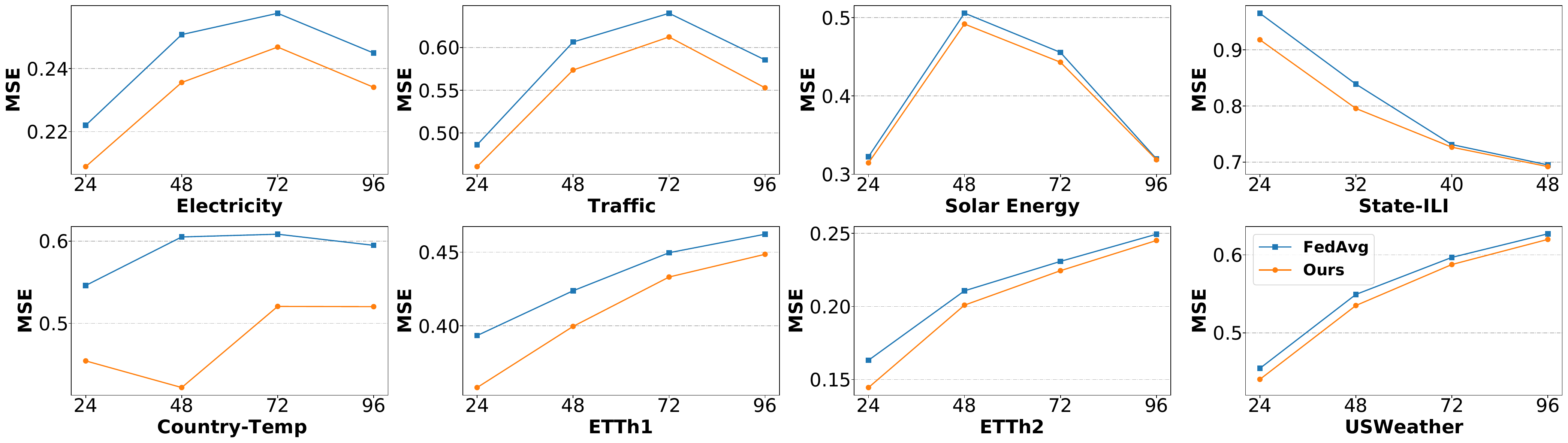} 
    \caption{The performance trend with time series data length.}
    \label{fig_rq5_length}
\end{figure*}
To simplify the investigation cases, we assume that input and output data lengths $L_{x}$ and $L_{y}$ are the same, and then, we change the data length from the default value $24$ to $96$.
Note that due to the limited number of timesteps in the State-ILI dataset (only $345$ in total, as shown in Table~\ref{tb_statistics}), we only examine data lengths from $24$ to $48$ for this dataset.

Intuitively, increasing the data length should decrease model performance since longer forecasting horizons are more challenging. This trend is observed in the Country-Temp, ETTh1, ETTh2, and USWeather datasets. However, in the Electricity, Traffic, and Solar Energy datasets, increasing the data length initially worsens performance, but further increases in data length help mitigate this decline. Interestingly, in the State-ILI dataset, a longer data length actually improves model forecasting. This may be because, while a longer output data length increases forecasting difficulty, a longer input data length provides more valuable contextual information. For example, in real-world scenarios, illness statistics like those in the State-ILI dataset exhibit seasonality, so having a longer observed data window is beneficial for accurate predictions.

Overall, in all cases, \modelname consistently outperforms its corresponding base federated learning framework, FedAvg, by a significant margin, demonstrating the robustness and effectiveness of our method across different data length settings.

\subsubsection{The Impact of Data Construction Interval}\label{sec_exp_frequency}
\begin{figure}[htbp]
    \centering
    \includegraphics[width=1.\columnwidth]{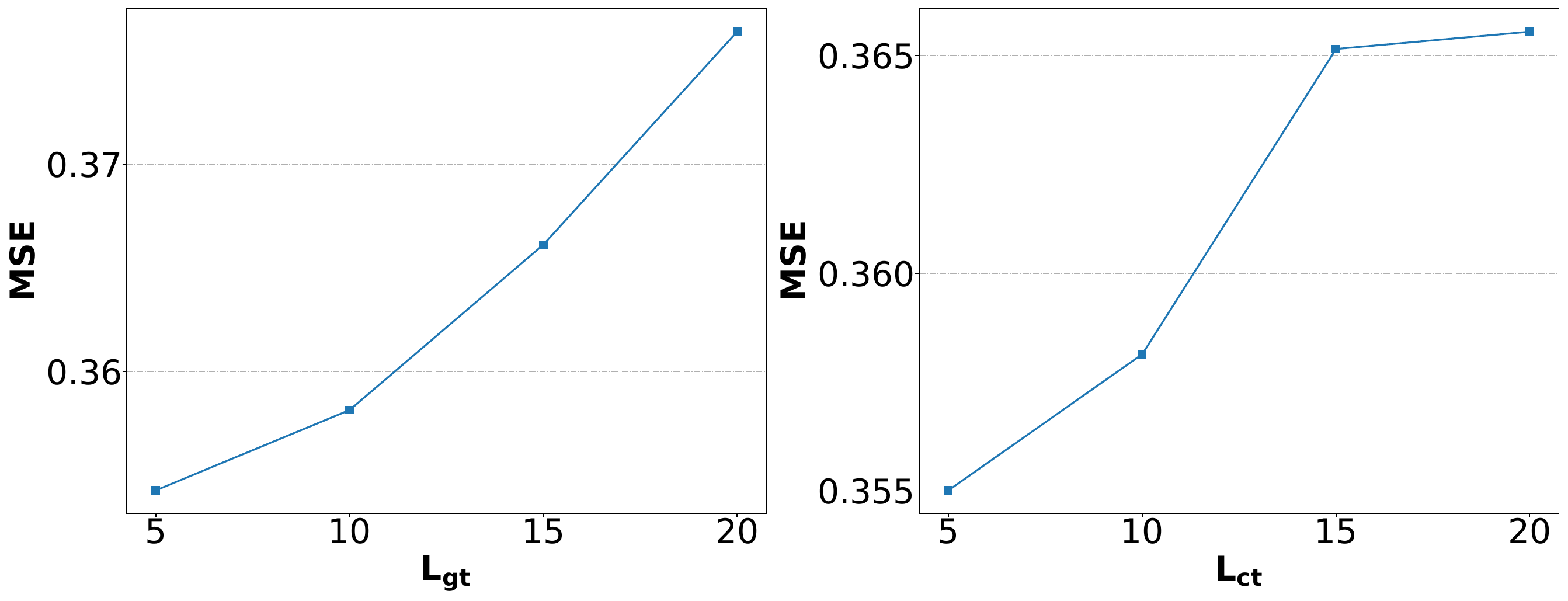} 
    \caption{The performance trend with the synthetic data construction frequency $L_{gt}$ and $L_{ct}$ on ETTh1. Similar trend can be observed on other datasets.}
    \label{fig_rq5_frequency}
\end{figure}
Figure~\ref{fig_rq5_frequency} illustrates the impact of synthetic data construction frequency on system performance. Due to space limitations, we present the results for the ETTh1 dataset, but similar trends are observed across the other seven datasets. The results indicate that smaller values of $L_{gt}$ and $L_{ct}$ lead to better performance. This is because more frequent construction of synthetic datasets allows them to quickly adapt to recent dynamics, thereby incorporating the latest knowledge from model updates. However, frequently updating synthetic datasets also increases computational costs, creating a trade-off between effectiveness and efficiency. In our experiments, we found that setting $L_{gt}$ and $L_{ct}$ to $10$ strikes a good balance.

\subsection{Further Study with Privacy Protection Mechanism (RQ5)}\label{sec_rq5_privacy}
\begin{figure}[htbp]
    \centering
    \includegraphics[width=1.\columnwidth]{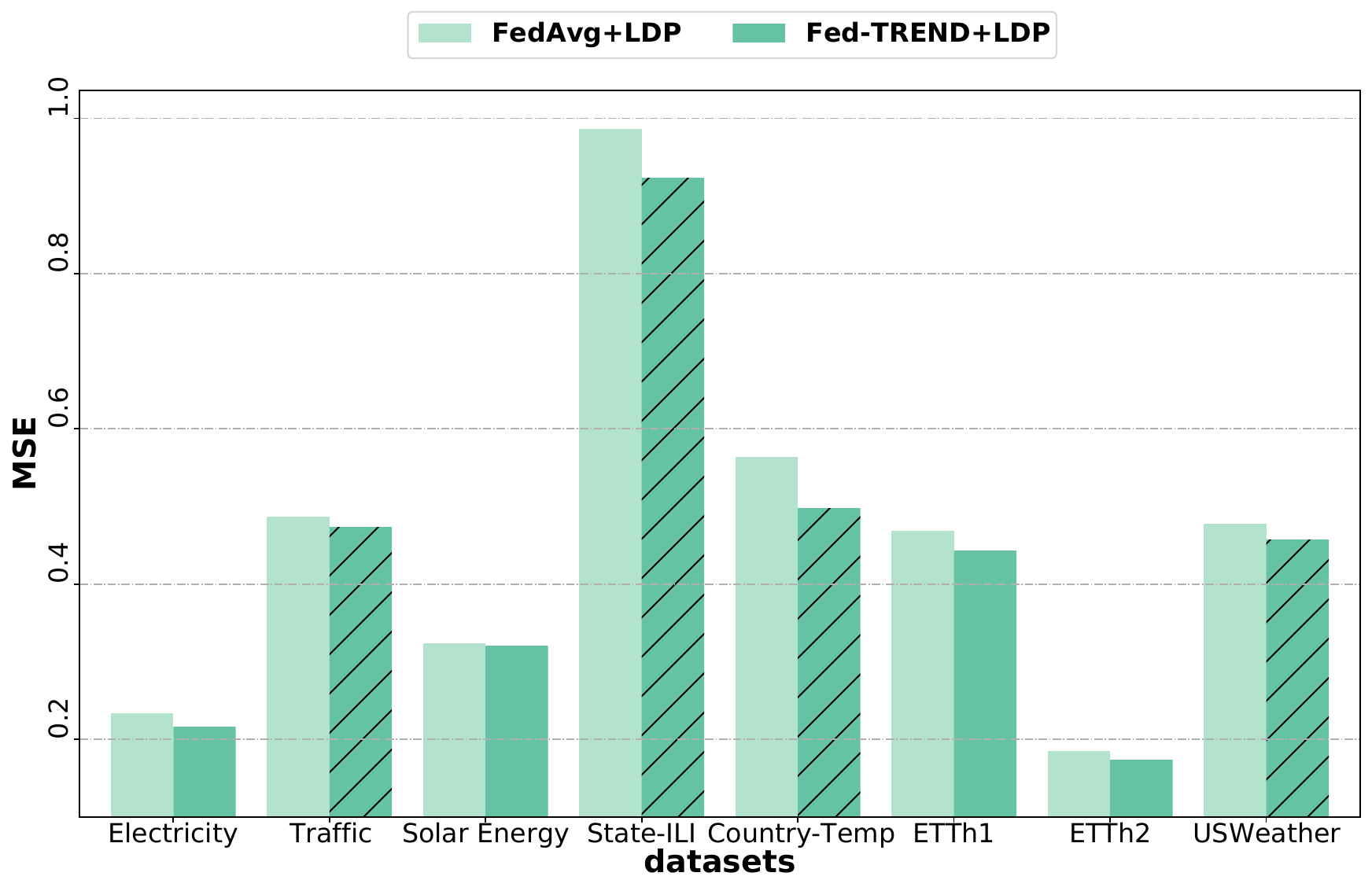} 
    \caption{The performance comparison of the base federated learning and the equipping of \modelname under the context of differential privacy.}
    \label{fig_dp}
\end{figure}
To enhance privacy protection, federated learning often incorporates privacy mechanisms. Among these, local differential privacy (LDP) is considered the gold standard and the most widely used approach~\cite{zhang2023comprehensive}. In this section, we evaluate whether \modelname can still improve the performance of baseline federated learning when using LDP. Specifically, we implement LDP with the Laplace mechanism by adding noise sampled from $\mathcal{N}(0, \lambda^{2}\mathbf{I})$ to the model parameters, where $\mathcal{N}$ represents the normal distribution, and we set $\lambda = 0.001$ to balance the trade-off between performance and privacy. As shown in Figure~\ref{fig_dp}, integrating \modelname with ``FedAvg + LDP" results in lower MSE scores across all datasets, indicating that \modelname remains effective in the context of differential privacy.

\section{Conclusion}\label{sec_conclusion}
This paper introduces \modelname, a federated time series forecasting framework designed to close the performance gap between federated and centralized time series forecasting by enhancing learning on heterogeneous data. 
Specifically, \modelname constructs two types of synthetic datasets based on clients' uploaded models and the aggregated global model to improve the consensus during clients' local training and to refine the global model aggregation, respectively. Since the synthetic data construction process does not require any prior knowledge and is performed on the central server, \modelname can be easily integrated with most federated learning frameworks without imposing a heavy computational burden on clients. Extensive experiments conducted on eight time series datasets using four popular forecasting models demonstrate the effectiveness and generalization capabilities of the proposed \modelname.

\bibliographystyle{IEEEtran}
\bibliography{IEEEexample}

\end{document}